\documentclass[twocolumn]{applemlr}

\usepackage{amsmath}
\usepackage{enumerate}
\usepackage{algorithm}
\usepackage{algpseudocode}
\usepackage{amsfonts}
\usepackage{amsthm}
\usepackage{cleveref}
\usepackage{diagbox}
\usepackage{colortbl}
\usepackage{amssymb}
\usepackage{xspace}
\usepackage{wrapfig}
\usepackage{adjustbox}
\usepackage{tabularx}
\usepackage{booktabs}
\usepackage{mathtools}
\usepackage{tikz}
\usepackage{enumitem}
\usepackage{silence}
\usepackage{dsfont}
\usepackage[table]{xcolor}
\usepackage[dvipsnames]{xcolor}
\usepackage{multirow}
\usepackage{makecell}
\usepackage{xfakebold}

\usepackage{amsmath,amsfonts,bm}

\def\eqref#1{equation~\ref{#1}}

\def\1{\bm{1}}

\DeclareMathAlphabet{\mathsfit}{\encodingdefault}{\sfdefault}{m}{sl}
\SetMathAlphabet{\mathsfit}{bold}{\encodingdefault}{\sfdefault}{bx}{n}

\definecolor{textgray}{HTML}{6E6E73}
\usetikzlibrary{positioning, calc}
\usetikzlibrary{decorations.pathmorphing}

\makeatletter
\patchcmd{\wrong@fontshape}{\@gobbletwo}{}{}{}
\makeatother
\numberwithin{equation}{section}
\makeatletter
\AtBeginDocument{
  \urlstyle{sf}
  
}
\makeatother

\definecolor{light}{RGB}{125, 125, 125}
\crefname{tcb@cnt@pbox}{code}{code}
\Crefname{tcb@cnt@pbox}{Code}{Code}
\crefname{assumption}{assumption}{assumption}
\Crefname{assumption}{Assumption}{Assumptions}

\newtcolorbox[auto counter]{pbox}[2][]{
  colback=white,
  title=Code~\thetcbcounter: #2,
  #1,fonttitle=\sffamily,
  fontupper=\sffamily,
  arc=2pt,
  colframe=bgcolor,
  coltitle=fgcolor,
  colbacktitle=bgcolor,
  toptitle=0.25cm,
  bottomtitle=0.125cm
}

\makeatletter
\newcommand\applefootnote[1]{%
  \begingroup
  \renewcommand\thefootnote{}%
  \renewcommand\@makefntext[1]{\noindent##1}%
  \footnote{#1}%
  \addtocounter{footnote}{-1}%
  \endgroup
}
\makeatother

\definecolor{cverbbg}{gray}{0.90}

\input{glyphtounicode}\pdfgentounicode=1
\pdfinterwordspaceon

\usepackage[T1]{fontenc}
\usepackage[utf8]{inputenc}
\usepackage{microtype}
\usepackage{graphicx}
\usepackage{placeins}
\usepackage{subcaption}
\usepackage{longtable}
\usepackage{listings}

\lstdefinelanguage{json}{
    basicstyle=\footnotesize\ttfamily,
    numbers=left,
    numberstyle=\tiny,
    stepnumber=1,
    numbersep=5pt,
    showstringspaces=false,
    breaklines=true,
    frame=single,
    backgroundcolor=\color{gray!5},
    string=[s]{"}{"},
    stringstyle=\color{blue!80!black},
    comment=[l]{:},
    commentstyle=\color{black},
}

\newcommand{\kalpha}{Krippendorff's $\alpha$}

\newcommand{\gptoss}{GPT-OSS-120B}
\newcommand{\gemini}{Gemini 2.5 Flash}
\newcommand{\deepseek}{DeepSeek-V3.1}
\newcommand{\qwen}{Qwen3.5-397B}

\title{Large Language Models for Machine Translation Quality Annotation: \\
Humans and Models Are Both Challenged}

\author{Hala Almaghout}
\author{Christian Federmann}
\author{Qin Gao}

\affiliation{Apple}

\abstract{
Large Language Models (LLMs) are considered to be a more efficient and cost-effective alternative to human judgment for Machine Translation (MT) evaluation. With MT
  evaluation spanning a large number of language pairs, domains and levels of
  annotation granularity, LLMs must be thoroughly evaluated across these dimensions before being reliably used as alternatives to human evaluation. In this paper, we evaluate the performance of LLMs for two prominent MT quality evaluation schemes: Multidimensional Quality Metrics (MQM) and Error Span Annotation (ESA) by comparing their agreement with human annotators. We present results on a long-context test set of 70 language pairs and the publicly available WMT23 and WMT25 data, investigating both score and error span annotation agreement across a variety of language pairs and domains. Our results show that while LLM agreement with human annotators exceeds agreement between human annotators for some evaluation tasks, both vary substantially across annotation schemes, language pairs and domains and remain unreliable for most settings. Furthermore, we identify challenges facing both human and LLM annotators: humans are particularly challenged by fine-grained MQM annotations and low-resource language pairs, while LLMs struggle with minor errors, wrong language variants and error span annotation. Our results highlight the potential for improvement for both human and LLM annotation performance, possibly through human-LLM collaborative annotation pipelines that address the reliability issues identified in this work.}

\metadata[Correspondence]{\sffamily \url{halmaghout@apple.com}, \url{chrf@apple.com}, \url{qingao@apple.com}}
\date{\sffamily October 7, 2026}

\begin{document}

\maketitle

\section{Introduction}

Translation evaluation is a challenging task, which requires evaluating multiple quality dimensions simultaneously. It includes judging the faithful transfer of meaning from the source language to the target language, the naturalness and fluency of the translated text in addition to its adherence to the target language linguistic and cultural structures, and the usability of the translation in a given context. Human evaluation is the gold standard to evaluate MT models. However, human evaluation is often expensive and time-consuming, slowing down MT model development. Furthermore, human evaluation suffers from quality and reliability issues. In addition to the highly subjective nature of translation evaluation, human evaluators differ in their skills, training, motivation and leniency. Annotation tasks also differ in their design, guidelines, compensation and nature of evaluated text. All these factors contribute to the unreliability of human evaluation. 

Several approaches aim to improve the reliability of human evaluation. This includes aggregating multiple annotations from different annotators for the same source and translation texts. Other work relies on more fine-grained evaluation approaches that ground evaluation rating in detailed error annotations such as MQM~\citep{lommel-etal-2014} and ESA~\citep{kocmi-etal-2024-error} instead of providing a single quality rating. While these fine-grained MT evaluation approaches offer a more principled way for human evaluation which improves their quality, interpretability and consistency, they are more time-consuming and expensive to obtain.  

Automatic MT evaluation metrics offer a solution to approximate human evaluation in a cost and time efficient way. Early automatic metrics such as BLEU~\citep{Papineni2002} and chrF~\citep{popovic-2015-chrf} rely on lexical matching between the translation and a human-produced reference translation. While being efficient to compute, they suffer from a severe limitation of penalizing correct translations which deviate from the reference wording.

With the advent of large pretrained language models, neural evaluation metrics emerged such as BLEURT~\citep{Sellam2020}, COMET~\citep{Rei2020} and MetricX~\citep{Juraska2024metricx}, offering the ability to represent and compare text beyond the surface level, which was the main limitation of surface-based automatic MT metrics. Neural MT evaluation metrics are built by fine-tuning language models on human evaluation data, with some offering error span annotation similar to MQM such as in xCOMET~\citep{Guerreiro2023xcomet}. While these model-based metrics often achieve better correlation with human judgment than surface-based metrics, they suffer from multiple limitations, mainly limited generalization ability on unseen language pairs and domains and lack of interpretability.       

With LLMs showing more powerful abilities, researchers began exploring whether these models might provide a cost and time effective alternative for human evaluation, allowing for faster model iterations. LLMs have been probed for performing automatic evaluation for a range of different tasks out-of-the-box without the need for training them on human judgment data. They also offer the ability to customize the evaluation process through prompting and providing explanation. In the MT domain, \citet{Kocmi2023gemba} introduced GEMBA, demonstrating that GPT-3.5 and GPT-4 prompted with simple, language-agnostic templates achieve state-of-the-art correlation with MQM human judgments on WMT22 data on the system level--- without any task-specific fine-tuning.

In this work, we evaluate LLMs for MT evaluation following two popular MT evaluation schemes: MQM and ESA. MQM involves fine-grained error annotations where the quality score is automatically calculated out of error categories and severities whereas ESA provides simpler error annotations with only severity and the score is input directly by the annotators instead of being calculated. Evaluating on these two different annotation schemes allows us to assess LLM performance across different levels of annotation granularity. We present results on a long-context test set of 70 language pairs and the publicly available WMT23 and WMT25 data, investigating both score and error span annotation agreement across a variety of language pairs and domains. We identify challenges facing both human and LLM annotators for fine-grained MT evaluation. Our analysis highlights the following main findings: 

\begin{itemize}
    \item LLM agreement with human annotators is indistinguishable from agreement between human annotators for some evaluation tasks.
    \item Agreement between both LLM and human annotators varies substantially between annotation schemes, language pairs and domains.
    \item Agreement between human annotators is severely impacted when performing fine-grained error annotations in MQM and for low-resource language pairs. 
    \item LLMs have particular challenges in identifying minor errors, flagging translations in the wrong language or variant and error span annotation, especially for rarer error categories. 
    \item Low agreement levels highlight the need for improving both human and LLM evaluation reliability. 
\end{itemize}

\section{Related Work}

\subsection{Automatic MT Evaluation Metrics}

The latest WMT25 metrics shared task~\citep{wmt25-metrics-findings} included tasks for both score prediction and span-level error detection. The evaluation results showed that LLM-based
metrics are the best performing metrics at the system level however they are outperformed by surface and embedding-based similarity metrics at the segment level. The task also demonstrated that
span-level error detection remains the hardest challenge, with the best autoraters (all LLM-based)
achieving only 9--13\% character-level F1 compared to 47\% for human annotators, which is consistent with our results. 

\citet{lrm-mt-eval-2025} improved alignment of Large Reasoning Models with human judgment by post training them on synthetic data that emulates human evaluation process. They identify minor errors
as being more misaligned with human annotations than major errors --- consistent with our findings.

\subsection{AI-Assisted and Collaborative Human Annotation}

\citet{kocmi-etal-2024-esaai} showed that AI-assisted human ESA annotation increases error recall
(1.6 spans per segment vs.\ 0.5 unassisted), improves inter-annotator consistency,
and reduces annotation budget by 24\% through pre-filtering of error-free segments. \citet{mqm-reannotation-2023} demonstrated that human revision of automatic MQM annotations
(GEMBA-MQM, AutoMQM) rather than annotating from scratch surfaces missed errors,
improves annotation quality without additional cost, and reduces annotation time
by up to 12\%.

\citet{sxs-mqm-2024} confirmed that performing MQM annotations for translation pairs simultaneously
yields higher inter-annotator agreement on the side-by-side comparison results, better error span consistency and more reliability in identifying equal-quality translations than
standard independent MQM annotations.

\subsection{Reliability of LLMs as Annotators}

\citet{bavaresco-etal-2024-llms} evaluated 11 LLMs on a 20-task NLP benchmark. They found that no single model consistently excels, LLMs correlate more with non-expert
than expert annotations, and chain-of-thought and few-shot prompting yield
inconsistent gains. They conclude that LLMs must be carefully validated before
deployment as evaluators -- a finding which our results also support.

\citet{calderon-etal-2025-alternative} propose a leave-one-out statistical test for
whether an LLM can replace a human annotator. We follow a similar leave-one-out approach in our experiments with the difference that \citet{calderon-etal-2025-alternative} calculate alignment on the instance level while we calculate alignment on the test set level using \kalpha. They find that success of the test is highly
task and annotation aspect-dependent and that low human inter annotator agreement (IAA) makes the test unreliable — a caveat applicable to the results
in our study.

\citet{kasner-etal-2025-span} evaluate LLMs on ESA error span annotation on a subset of the WMT24 human evaluation data. They found that human-human IAA consistently exceeds
LLM-human IAA for span annotation, and that LLM-generated character-level span
indices require post-hoc re-alignment --- a challenge we independently observe and
address. 

Previous work evaluated LLMs against human annotators for machine translation quality annotation focusing on either MQM or ESA, and it was limited to a small number of language pairs. By contrast, our work compares the agreement of LLMs versus human annotators on both ESA and MQM, exploring both score and span annotation for both schemes. Our MQM results cover the full MQM taxonomy on 70 language pairs and our ESA results cover multiple language pairs and domains, questioning the reliability of the human baseline across these dimensions. We present results using two different annotator pools and annotation platforms and supported by statistical significance tests.

\section{Experimental Setup}
\subsection{Data}

We present our findings on three data sets:

  \begin{itemize}
      
      \item \textbf{WMT23 MQM.} The MQM annotations from the
      WMT23 Metrics Shared Task~\citep{Freitag2023wmt23metrics} for two
      language pairs: English-to-German (en-de) and Chinese-to-English (zh-en), which include three independent annotation rounds per segment.

      \item \textbf{Long-Context Web MQM.} Our internal test set consisting
      of multi-paragraph text extracted from crawled web pages. Target translations
      were produced by human translators and a variety of MT systems, including
      neural and LLM-based models.

      \item \textbf{WMT25 ESA.} The Error Span Annotation
      (ESA)~\citep{kocmi-etal-2024-error} data from the WMT25 Human Evaluation
      Data~\citep{kocmi-etal-2025-findings}, covering 14 language pairs across 5 domains\footnote{Due to access restrictions imposed by the YouTube platform, WMT25 speech domain was not included in the analysis and represents a limitation of the current work.}.
  \end{itemize}

  For our MQM annotations on the Long-Context Web MQM test set, professional annotators followed guidelines adapted
  from~\citet{freitag-etal-2021-experts}, selecting error spans in the source
  and/or target text and specifying the category and severity of each error
  using a specially designed interface. We conducted two annotation rounds covering all segments, yielding two
  independent annotations per sample from different annotators. Thus, our work presents results under two different annotation settings with different annotator pools and annotation platforms. Appendix~\ref{sect:lct_human_annotation} provides more details about the human annotation process for the Long-Context Web MQM test set.
  Tables~\ref{tab:lct_stats} and~\ref{tab:wmt23_mqm_stats} show data statistics
  for the Long-Context Web MQM and WMT23 MQM test sets, respectively. All data sets have segments which are predominantly long-context.

\subsection{Large Language Models}

We use the following LLMs for our evaluation: Gemini 2.5 Flash~\citep{comanici2025gemini25pushingfrontier}, \gptoss{}~\citep{openai2025gptoss120bgptoss20bmodel}, \deepseek{}~\citep{deepseekai2024deepseekv3technicalreport} and \qwen{}~
\citep{qwen3.5}. All inference is performed with $\texttt{temperature}=0$ and $\texttt{top\_p}=1$. Open-weight models are self-hosted while \gemini{} is run through Vertex AI in April--May 2026. 

The models receive an adapted version of the annotation
  guidelines provided to human annotators (refer to Figure~\ref{fig:mqm_prompt} for MQM and Figure~\ref{fig:esa_prompt} for ESA). For WMT23, we use the original MQM error annotation instructions from~\citet{freitag-etal-2021-experts}, which define the error taxonomy used in the WMT23 annotations.
  
 We report results using zero-shot prompting, as adding few-shot examples
  did not improve LLM-human agreement. We disable reasoning for all open-weight models except for \gptoss{} which was run with \texttt{reasoning\_effort}$=$\texttt{low}. The output token limit is set for open-weight models per task: 4096 tokens for WMT23 MQM and WMT25 ESA, and 8192 for Long-Context Web MQM, raised to 16384 for \deepseek{}. No explicit output-token limit or reasoning effort is set for \gemini{}, which therefore used the serving API's default. We use guided decoding to make the LLM response adhere to the specified output format. Because every model is evaluated on the same segments, we exclude any segment for which at least one LLM failed to return a parseable annotation, removing 1.1\% of WMT23 MQM, 5.1\% of Long-Context Web MQM, and 7.3\% of WMT25 ESA segments.

  A critical issue emerged when calculating span annotation agreement with LLM-provided span indices: at least 80\% of extracted
  spans based on these indices did not match the corresponding span text in
  the LLM output for all LLMs on MQM and ESA data sets. To correct this, we re-aligned spans by searching for the
  LLM-provided span text within the source (for source errors) or target
  (for target errors) and recording the first match start and end indices. This
  post-processing step was necessary to enable accurate error localization
  for annotation comparison.

\subsection{MQM Scoring}

MQM scores are calculated out of the error annotations provided by both humans and the LLM using the standard MQM scoring formula~\citep{freitag-etal-2021-experts} which weighs major errors by 5, minor errors by 1 and punctuation errors by 0.1.

\subsection{Annotation Evaluation}

 We evaluate LLM annotations across the following aspects:
  scores, error counts, and span-level annotations. We do so by comparing LLM annotation agreement with human annotations, treating each human annotation
  round as an independent annotator.

\textbf{Score and Error Count Evaluation.} We use \kalpha{}~\citep{krippendorff04} with interval distance to measure inter-annotator agreement (IAA) on segment-level MQM and ESA scores as well as major and minor error counts per segment for MQM. \kalpha{} is a widely-used agreement coefficient that provides several advantages over
  correlation measures such as Pearson and Spearman: it accounts for chance agreement, handles multiple annotations and measures agreement on absolute scores rather than merely relative co-variation — enabling assessment of how
  consistently annotators agree on absolute translation quality. We compute LLM-human IAA using a leave-one-out approach: each human annotation round is in turn excluded and replaced by the LLM annotation, and IAA is computed within the resulting annotator group. The reported  
  LLM-human IAA is the average across all such groups.

  \textbf{Span Annotation Evaluation.} We compute the following F1-based metrics to
  evaluate error span identification and localization:

  \begin{itemize}
     \item \textbf{Character overlap:} Measures character-level overlap
      between predicted and gold spans. Each predicted span is matched to
      the gold span with maximum character overlap (category-agnostic),
      and F1 is computed based on the number of overlapping characters.

      \item \textbf{Span overlap:} Measures F1 for predicting
      spans that overlap with gold spans regardless of category and severity (minimum one character overlap required).
     
      \item \textbf{Category/severity-specific span overlap:} Same as span overlap but with the requirement of having the same error category/severity when matching with gold spans.

      \item \textbf{No error:} F1 for correctly identifying error-free segments. 
  \end{itemize}

F1 metrics are computed for all human-human and human-LLM annotation pairs and averaged within each group.

\subsection{Significance Testing}

We tested the statistical significance of the hypothesis of LLM-human agreement being equal to the human-human agreement in our results. We used two-sided paired bootstrap resampling with replacement on samples clustered by source segments with 10,000 bootstrap iterations and treating a difference as significant at $p<0.05$. LLM-human and human-human agreement are calculated on the same resampled source segments for each iteration.  For Tables~\ref{tab:esa_llm_exceeds_hh} and~\ref{tab:mqm_llm_exceeds_hh} comparing agreement across multiple language pairs within different domains or language pair directions, we used the Benjamini–Hochberg false discovery rate (FDR) procedure \citep{benjamini-hochberg-1995} separately to each LLM within each table to correct the p-values applied, controlling the FDR rate at 0.05.

\section{Results}

\subsection{MQM}
\label{sect:WMT23_MQM}
\begin{figure}[ht]
\centering
\includegraphics[width=\columnwidth]{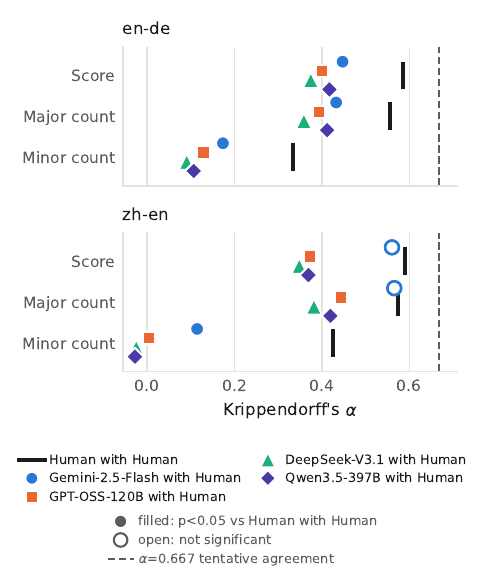}
\caption{Inter-annotator agreement on the WMT23 MQM test set, showing Krippendorff's $\alpha$ for MQM scores, major error counts, and minor error counts.}
\label{fig:alpha_by_language_pair_average_MQM_wmt23}
\end{figure}

\subsubsection{Score and Error Count Evaluation}

 Figure~\ref{fig:alpha_by_language_pair_average_MQM_wmt23} shows agreement for each of MQM score,  major error counts and minor error counts on the WMT23 test set.  For MQM score, human-human agreement is significantly better than LLM-human agreement for both en-de and zh-en. The only exception is \gemini{}-human MQM score agreement for zh-en which is on par with human-human agreement.  Overall, agreement between different annotator combinations is well below the 0.667 threshold for tentative agreement. 

Given that MQM scores are derived from major and minor error counts, we conduct agreement experiments for segment-level major and minor error counts for different annotator pairs to better understand how each error type contributes to the overall score agreement. For major errors, \gemini{}-human agreement approaches human-human agreement for zh-en but falls behind it for en-de while other models' agreement is further behind. For minor errors, human-human agreement substantially exceeds LLM-human agreement for both language pairs, especially for zh-en where the margin is larger. This demonstrates that LLMs have more difficulty in identifying minor errors that involve subtle quality distinctions. The difficulty of minor error annotation is particularly consequential for MT evaluation: as MT system quality improves, minor errors increasingly become the primary differentiating factor
  among closely ranked systems, making reliable minor error annotation essential for meaningful quality estimation.

These findings are consistent with results on the Long-Context MQM test set with wider language pair coverage (Appendix~\ref{sect:mqm_lct_results}), which was independently annotated using different annotator pools and annotation platform.

\subsubsection{Span Annotation Evaluation}
\label{sect:WMT23_MQM_error_spans}

\begin{table}[t]
\centering
\footnotesize
\setlength{\tabcolsep}{2pt}
\begin{tabular}{lrrrrr}
\toprule
\textbf{Error category} & \textbf{H-H} & \textbf{Gem} & \textbf{GPT} & \textbf{DS} & \textbf{Qwen} \\
\midrule
Span overlap & .51\phantom{$^{*}$} & .43$^{*}$ & .44$^{*}$ & .39$^{*}$ & .39$^{*}$ \\
Character overlap & .43\phantom{$^{*}$} & .39$^{*}$ & .31$^{*}$ & .27$^{*}$ & .35$^{*}$ \\
Severity minor & .33\phantom{$^{*}$} & .17$^{*}$ & .16$^{*}$ & .13$^{*}$ & .12$^{*}$ \\
Mistranslation & .37\phantom{$^{*}$} & .35$^{*}$ & .37\phantom{$^{*}$} & .32$^{*}$ & .34$^{*}$ \\
Severity major & .45\phantom{$^{*}$} & .43$^{*}$ & .43$^{*}$ & .31$^{*}$ & .41$^{*}$ \\
Awkward & .19\phantom{$^{*}$} & .13$^{*}$ & .12$^{*}$ & .01$^{*}$ & .02$^{*}$ \\
Grammar & .18\phantom{$^{*}$} & .19\phantom{$^{*}$} & .10$^{*}$ & .13$^{*}$ & .16$^{*}$ \\
Punctuation & .28\phantom{$^{*}$} & .04$^{*}$ & .06$^{*}$ & .02$^{*}$ & .03$^{*}$ \\
Inappropriate for context & .07\phantom{$^{*}$} & .05\phantom{$^{*}$} & .05\phantom{$^{*}$} & .05\phantom{$^{*}$} & .06\phantom{$^{*}$} \\
Omission & .36\phantom{$^{*}$} & .30$^{*}$ & .14$^{*}$ & .06$^{*}$ & .15$^{*}$ \\
Spelling & .14\phantom{$^{*}$} & .05$^{*}$ & .04$^{*}$ & .01$^{*}$ & .06$^{*}$ \\
Addition & .26\phantom{$^{*}$} & .21$^{*}$ & .11$^{*}$ & .11$^{*}$ & .16$^{*}$ \\
Source error & .11\phantom{$^{*}$} & .00$^{*}$ & .00$^{*}$ & .00$^{*}$ & .00$^{*}$ \\
Register & .00\phantom{$^{*}$} & .02$^{*}$ & .01$^{*}$ & .01\phantom{$^{*}$} & .01$^{*}$ \\
Character encoding & .28\phantom{$^{*}$} & .02$^{*}$ & .01$^{*}$ & .33\phantom{$^{*}$} & .06$^{*}$ \\
Inconsistency & .08\phantom{$^{*}$} & .04$^{*}$ & .03$^{*}$ & .01$^{*}$ & .05\phantom{$^{*}$} \\
Untranslated text & .12\phantom{$^{*}$} & .00$^{*}$ & .00$^{*}$ & .00$^{*}$ & .00$^{*}$ \\
No error & .10\phantom{$^{*}$} & .12\phantom{$^{*}$} & .07$^{*}$ & .00$^{*}$ & .15$^{*}$ \\
Other & .02\phantom{$^{*}$} & .00$^{*}$ & .00$^{*}$ & .00$^{*}$ & .00$^{*}$ \\
Non translation & .17\phantom{$^{*}$} & .00$^{*}$ & .00$^{*}$ & .00$^{*}$ & .00$^{*}$ \\
Date format & .10\phantom{$^{*}$} & .09\phantom{$^{*}$} & .00$^{*}$ & .05\phantom{$^{*}$} & .10\phantom{$^{*}$} \\
Inconsistent use & .01\phantom{$^{*}$} & .00\phantom{$^{*}$} & .00\phantom{$^{*}$} & .00\phantom{$^{*}$} & .00\phantom{$^{*}$} \\
Text Breaking & .02\phantom{$^{*}$} & .00\phantom{$^{*}$} & .00\phantom{$^{*}$} & .00\phantom{$^{*}$} & .00\phantom{$^{*}$} \\
Name format & .04\phantom{$^{*}$} & .04\phantom{$^{*}$} & .00$^{*}$ & .00$^{*}$ & .00$^{*}$ \\
Currency format & .04\phantom{$^{*}$} & .00\phantom{$^{*}$} & .00\phantom{$^{*}$} & .00\phantom{$^{*}$} & .00\phantom{$^{*}$} \\
\bottomrule
\end{tabular}
\caption{Span-annotation F1 by error category on the WMT23 MQM test set, averaged over language pairs and ordered by decreasing category frequency. H-H is agreement between human annotation rounds; the remaining columns are each LLM's agreement with the human rounds. Gem = Gemini 2.5 Flash, GPT = GPT-OSS-120B, DS = DeepSeek-V3.1, Qwen = Qwen3.5-397B. $^{*}$ marks a difference from H-H that is statistically significant at $p<0.05$. Categories whose human-human baseline rests on fewer than 5 source segments are omitted, since no comparison against it is estimable: Address format, Time format.}
\label{tab:f1_by_categories_MQM_wmt23}
\end{table}

Table~\ref{tab:f1_by_categories_MQM_wmt23} shows span annotation evaluation on the WMT23 test set in terms of F1 score, averaged across all language pairs. Agreement levels for high-frequency error categories are notably higher than low-frequency categories. In addition, human-human agreement exceeds LLM-human agreement for most span annotation evaluation metrics with some exceptions which include Mistranslation where \gptoss-human is on par with human-human agreement and Grammar where \gemini{}-human agreement is on par with human-human agreement. Human-human agreement on span overlap reaches at most F1 score = 0.51, indicating the inherent difficulty of error span annotation. Notably, the agreement levels for the lower-frequency error categories are extremely low for all annotator combinations, with many categories achieving zero F1 score by all LLMs such as Currency Format. This suggests that while LLMs show stronger agreement with human annotators for frequent error categories such as mistranslation, they particularly struggle to identify rarer error categories, possibly because these errors are underrepresented in the LLM training data. In terms of error severity, both LLM-human and human-human agreements are lower for minor errors than major errors, which confirms error count results in Section~\ref{sect:WMT23_MQM}. When it comes to identifying error-free segments (the No Error category), human-human agreement is significantly outperformed by \qwen{}-human agreement and is no different from \gemini{}-human agreement while both \deepseek{}-human and \gptoss{}-human agreements are significantly inferior to human-human agreement. However, all agreements remain at or below 0.15 in absolute terms. This likely stems from the long-context nature of the test segments and the high disagreement on annotating minor errors: in near-perfect translations spanning multiple paragraphs, annotators frequently disagree on whether subtle imperfections constitute minor errors, leading to inconsistent error-free classifications. Notably, \deepseek{}-human agreement on identifying error-free segments is very low across all test sets (Tables~\ref{tab:f1_by_high_freq_categories_ESA} and~\ref{tab:f1_by_categories_MQM}).
Span annotation results on the Long-Context MQM test set (Appendix~\ref{sect:MQM_error_spans}) mirror this pattern with human-human agreement exceeding LLM-human for most categories, and the gap is widening for the low-frequency ones.

\subsection{ESA}
\subsubsection{Score Evaluation}
\label{sect:ESA_scores}
\begin{figure*}[t]
\includegraphics[width=\textwidth]{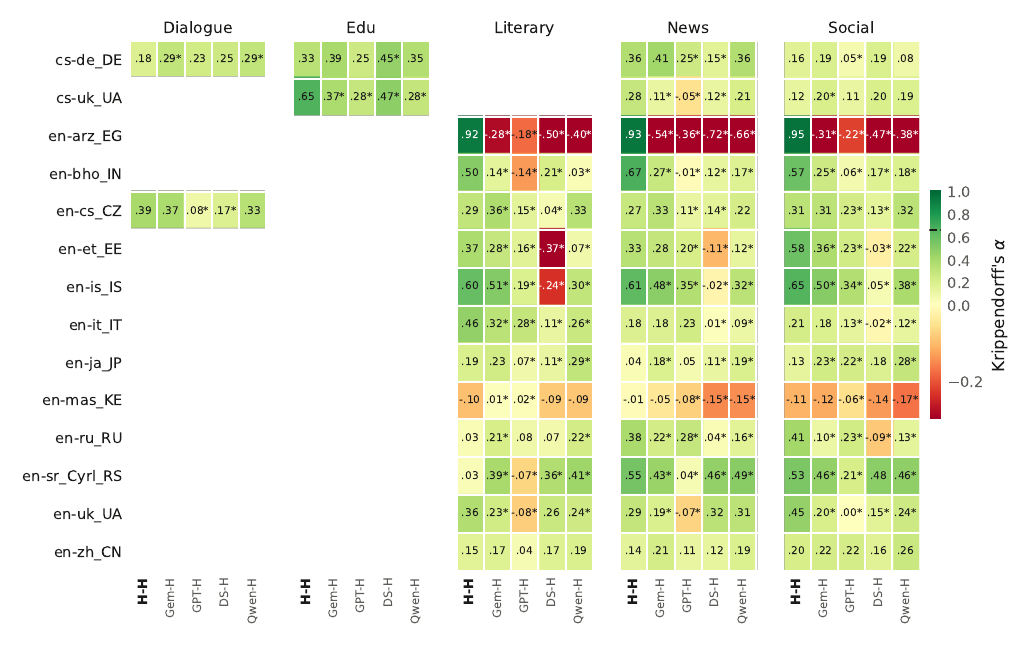}
\caption{Krippendorff's $\alpha$ agreement on ESA scores for each language pair and domain on the WMT25 ESA test set. Within each domain, the columns are H-H (Human with Human), Gem-H (Gemini-2.5-Flash with Human), GPT-H (GPT-OSS-120B with Human), DS-H (DeepSeek-V3.1 with Human), Qwen-H (Qwen3.5-397B with Human). The dashed line on the colorbar marks $\alpha = 0.667$. $*$ marks cells where LLM-human agreement differs significantly from the human-human agreement of the same cell (paired cluster bootstrap on source segments, $p<0.05$).}
\label{fig:esa_alpha_heatmap}
\end{figure*}

Figure~\ref{fig:esa_alpha_heatmap} presents ESA score agreement for each language pair and domain on the WMT25 ESA test set. Consistent with MQM score results, ESA scores exhibit agreement below the \kalpha{} threshold for tentative conclusions (0.667) for nearly all language pairs and domains. Exceptions include en-arz\_EG across all domains, though as discussed below, the high en-arz\_EG agreement reflects systematic issues rather than annotation quality.  Both human-human and LLM-human
  agreement vary substantially across domains within the same language pair,
  indicating that domain characteristics significantly influence annotation
  difficulty. For a few language pairs such as English-to-Japanese (en-ja\_JP), \qwen{}-human and \gemini{}-human agreement exceeds human-human agreement across all domains. Conversely, for language pairs such as English-to-Egyptian Arabic (en-arz\_EG) and English-to-Bhojpuri (en-bho\_IN), human-human agreement substantially outperforms LLM-human agreement across all domains. A notable result is English-to-Maasai (en-mas\_KE), which has a remarkably low agreement across all annotator pairs consistent with its status as a low-resource language pair. 

  A striking finding emerges for en-arz\_EG: LLM-human agreement shows negative \kalpha{} for LLMs which indicates systematic disagreement whereas human-human agreement is remarkably high (>0.90 for all domains), indicating systematic agreement. Manual examination of
  sample annotations reveals the root cause: Most translations (60\%) for en-arz\_EG are in the wrong Arabic variant (Modern Standard Arabic instead of Egyptian Arabic)\footnote{Percentage of en-arz\_EG translations in wrong variant are estimated by calculating percentage of translations which has a score of 0 by one of the human annotators and a single error span, which is how annotators are instructed to treat translations with wrong language according to the ESA annotation instructions in Figure~\ref{fig:esa_prompt}}. While humans consistently identified these samples and scored them as zero, the LLMs consistently assigned these translations a high score because they are accurate translations despite explicit instructions in the ESA prompt (Figure~\ref{fig:esa_prompt}) to score translation in the wrong language/variant by zero. This failure mode highlights a critical limitation in the LLMs' ability to distinguish between closely related language variants—a capability essential for locale-specific quality evaluation.

\begin{table}[ht]
\centering
\small
\resizebox{\columnwidth}{!}{
\begin{tabular}{lrcccc}
\toprule
\textbf{Domain} & \textbf{\#LPs} & \textbf{Gemini} & \textbf{GPT-OSS} & \textbf{DeepSeek} & \textbf{Qwen} \\
\midrule
Dialogue & 2 & 1/0 & 0/1 & 0/1 & 1/0 \\
Edu & 2 & 0/1 & 0/1 & 1/1 & 0/1 \\
Literary & 12 & 4/6 & 1/9 & 1/7 & 3/6 \\
News & 14 & 1/6 & 0/11 & 1/11 & 1/7 \\
Social & 14 & 2/7 & 2/10 & 0/8 & 1/9 \\
\midrule
All domains & 44 & 8/20 & 3/32 & 3/28 & 6/23 \\
\bottomrule
\end{tabular}}
\caption{Language pairs per domain whose LLM-human \kalpha{} agreement on ESA
scores differs significantly from human-human agreement, on the WMT25 ESA test set,
shown separately for each LLM.
The first number counts language pairs where LLM-human \kalpha{} significantly exceeds human-human agreement, the second those where human-human significantly exceeds LLM-human. LPs with no significant difference in either direction appear in neither number.
Gemini = Gemini 2.5 Flash, GPT-OSS = GPT-OSS-120B, DeepSeek = DeepSeek-V3.1, Qwen = Qwen3.5-397B. Benjamini--Hochberg FDR applied separately to each LLM's comparisons, $p<0.05$.}
\label{tab:esa_llm_exceeds_hh}
\end{table}

Table~\ref{tab:esa_llm_exceeds_hh} shows the number of language pairs per domain where LLM-human agreement significantly exceeds human-human agreement versus the number of language pairs where human-human agreement significantly exceeds LLM-human agreement for ESA scores on the WMT25 ESA test set. In contrast to MQM score agreement results, human-human ESA score agreement significantly exceeds LLM-human agreement for the majority of language pairs and domains for most LLMs. This might be due to the fact that ESA has a simpler annotation scheme where scores are input directly by annotators rather than calculated from error spans, which likely reduces disagreement whereas MQM calculates the scores by weighting the span counts which might elevate the variance. Comparing the LLM-human agreement between different models, \gemini{}-human agreement is higher than other models' agreement for most language pairs and domains. Notably, \gemini{}-human agreement significantly outperforms human-human agreement for 4 out of 12 language pairs in the Literary domain whereas it significantly underperforms human-human agreement for 6 out of 14 for the news domain and 7 out of 14 for the social domain.

\subsubsection{Span Annotation Evaluation}

Table~\ref{tab:f1_by_high_freq_categories_ESA} shows span annotation evaluation on the WMT25 ESA test set in terms of F1 score, averaged across all language pairs and domains. Unlike MQM, ESA has fewer error categories to measure agreement on. Similar to the MQM span annotation results in Section~\ref{sect:WMT23_MQM_error_spans}, human-human agreement is significantly better than LLM-human agreement for all span annotation evaluation metrics. In addition, both LLM-human and human-human agreements on major errors are higher than those of minor errors. Both humans and LLMs achieve low agreement on the missing span category. Notably, LLM-human agreement is on the same level for most categories with a few exceptions such as \deepseek{}-human agreement (0.05) being substantially lower than other models' agreement on identifying error-free segments.  

\begin{table}[t]
\centering
\footnotesize
\setlength{\tabcolsep}{2pt}
\ifx\iaahdrbox\undefined\newsavebox{\iaahdrbox}\fi
\sbox{\iaahdrbox}{\footnotesize\textbf{Qwen}}
\begin{tabular}{lrrrrr}
\toprule
\textbf{Span agreement} & \makebox[\wd\iaahdrbox][c]{\textbf{H-H}} & \makebox[\wd\iaahdrbox][c]{\textbf{Gem}} & \makebox[\wd\iaahdrbox][c]{\textbf{GPT}} & \makebox[\wd\iaahdrbox][c]{\textbf{DS}} & \makebox[\wd\iaahdrbox][c]{\textbf{Qwen}} \\
\midrule
Character overlap & .43\phantom{$^{*}$} & .17$^{*}$ & .16$^{*}$ & .17$^{*}$ & .18$^{*}$ \\
Span overlap & .45\phantom{$^{*}$} & .30$^{*}$ & .31$^{*}$ & .31$^{*}$ & .31$^{*}$ \\
Minor severity & .23\phantom{$^{*}$} & .15$^{*}$ & .12$^{*}$ & .16$^{*}$ & .15$^{*}$ \\
Major severity & .32\phantom{$^{*}$} & .22$^{*}$ & .21$^{*}$ & .21$^{*}$ & .22$^{*}$ \\
No error & .47\phantom{$^{*}$} & .37$^{*}$ & .25$^{*}$ & .05$^{*}$ & .29$^{*}$ \\
Missing span & .20\phantom{$^{*}$} & .15$^{*}$ & .09$^{*}$ & .07$^{*}$ & .11$^{*}$ \\
\bottomrule
\end{tabular}
\caption{Span-annotation F1 by span-agreement measure on the WMT25 ESA test set, averaged over language pairs and domains and ordered by decreasing frequency. H-H is agreement between human annotation rounds; the remaining columns are each LLM's agreement with the human rounds. Gem = Gemini 2.5 Flash, GPT = GPT-OSS-120B, DS = DeepSeek-V3.1, Qwen = Qwen3.5-397B. $^{*}$ marks a difference from H-H that is statistically significant at $p<0.05$.}
\label{tab:f1_by_high_freq_categories_ESA}
\end{table}

\subsection{Annotation Quality Validation}
\label{sect:annotation_quality_valid}
\begin{figure}[ht]
\centering
\includegraphics[width=\columnwidth]{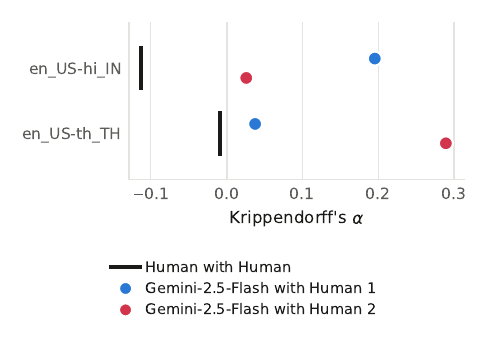}
\caption{Inter-annotator agreement for MQM scores on the Long-Context Web MQM test set for en\_US-hi\_IN and en\_US-th\_TH, showing Gemini 2.5 Flash's agreement with each individual human annotation round and agreement between human annotation rounds.}
\label{fig:alpha_by_language_pair_direction_en_X_score_gemini-2-5-flash_MQM}
\end{figure}

We observe that for many language pairs, there is a substantial difference in the LLM-human agreement across human annotation rounds. For example Figure~\ref{fig:alpha_by_language_pair_direction_en_X_score_gemini-2-5-flash_MQM} shows \gemini's agreement with each human annotation round on MQM scores for English-to-Hindi (en\_US-hi\_IN) and English-to-Thai (en\_US-th\_TH). For these language pairs the human-human agreement is negative or close to zero and \gemini's agreement with one of the human annotation rounds substantially exceeds its agreement with the other. This suggests that one of the two human annotation rounds has issues which cause inconsistencies for these language pairs and that LLMs can serve as diagnostic tools for flagging such problematic rounds for further investigation.

\section{Conclusion}

We evaluate LLMs as MQM and ESA annotators using inter-annotator agreement on a long-context test set of 70 language pairs and the publicly available WMT23 and WMT25 data, investigating both score and error span annotation agreement across a variety of language pairs, domains and annotation settings, using four LLMs: \gemini{}, \gptoss{}, \deepseek{} and \qwen{}. We demonstrate that LLM-human agreement exceeds human-human agreement for some language pairs and domains, which highlights the potential of using LLMs for machine translation quality estimation. However, our findings suggest that LLMs are not yet a reliable substitute for human MT evaluation. First, LLM-human agreement levels vary considerably between different language pairs and domains and between different LLMs. This highlights the need for conducting thorough LLM-human agreement tests with tight quality controls before using LLMs for MT evaluation in specific domains and language pairs. Second, and more importantly, agreement levels remain well below the \kalpha{} threshold for tentative conclusion agreement (0.667), particularly for low-resource language pairs, highlighting a substantial potential for enhancing both human and LLM annotation performance to achieve more reliable annotations. Following \citet{calderon-etal-2025-alternative}, we caution against taking LLM-human agreement superiority over human-human agreement as an indicator that LLMs can substitute humans in these settings, given the weak human reliability. 

Our results identify several key challenges facing both humans and LLMs for machine translation quality evaluation. LLM annotators have particularly low agreement with human annotators on minor errors and translations in the wrong language variant. Furthermore, LLMs struggle with error span annotation for all error categories, particularly rarer errors, suggesting that targeted prompt engineering or fine-tuning may offer a path to improvement. Human annotators are particularly challenged on fine-grained MQM error annotation and low-resource language pairs, which needs to be addressed through clearer guidelines, better annotator training and improved annotation processes that reduce annotator fatigue. We also show that LLMs may be used as diagnostic tools to flag inconsistencies in human annotations. Although the identified weaknesses could be addressed independently for LLMs and humans, we believe a combined LLM-human collaborative framework that leverages both complementary strengths would offer a more promising path to reliable and efficient MT evaluation.

\section{Limitations}

One limitation of this work is that the \kalpha{} threshold for reliable agreement is not well-established for the fine-grained MT quality annotations such as MQM. The current widely used 0.667 threshold in the research community for establishing tentative agreement might not be realistic for such a task with inherent difficulty and subjectivity. More research is needed to establish realistic annotator reliability thresholds under different settings such as text length, domain and number of annotations. Furthermore, more research is needed to identify the factors for low-agreement levels for both humans and LLMs,
  particularly for long-context documents, such as annotator fatigue for humans and attention dispersion for LLMs.

For licensing reasons we cannot publish our internally-annotated Long-Context Web MQM test sets, which presents a significant limitation of the replicability of the MQM results presented in this work. While we tried to cover a variety of language pairs, domains and annotation schemes in our work, more studies should be conducted to gauge the LLM ability for MT quality evaluation on larger data sets spanning a wider set of language pairs and domains. 

We presented our results using four LLMs, a single prompt in a zero-shot setting and a simple heuristic for span correction. With the growing variety of available LLMs, more evaluation needs to be conducted using different LLMs with ablation for different prompting strategies and a better approach for span localization.  

Data contamination is a concern with results run through modern LLMs. The cut-off dates for the models used in this paper (\gptoss{}: June 2024, \gemini{}: January 2025, \deepseek{}: July 2025, \qwen{}: estimated mid-to-late 2025) are likely before the release of the WMT25 ESA test set. However, the WMT23 MQM test set might be included in the training data of these models, which might inflate the LLM-human agreement. Our Long-Context Web MQM has never been publicly released so it cannot be in the training data of any of these models.   

Significance testing used in this paper examined the null hypothesis of human-human and LLM-human agreements being equal. A more refined statistical testing approach should in addition establish equivalence between human-human and LLM-human agreements. 

Using LLMs for evaluation poses the risk of amplifying biases in evaluated models. This was visible in our results in which LLMs failed to penalize translations in Modern Standard Arabic where Egyptian Arabic was expected, biasing towards the variant with more representation in the training data.  

\clearpage
\bibliographystyle{acl_natbib}
\bibliography{custom}

\clearpage
\FloatBarrier

\appendix
\section{Appendix}
\label{sec:appendix}

\subsection{Long-Context Web MQM Human Annotation}
\label{sect:lct_human_annotation}

The Long-Context Web MQM test set was constructed through an annotation vendor that handled annotator recruitment, annotation project management and quality control, and task-specific training. The vendor owns and maintains a secure annotation platform with no association with third-party providers or crowdsourcing. We provided the vendor with annotation guidelines. We also provided the vendor with annotation UI mock-ups to build the annotation UI. We kept direct communication with the vendor to address annotation issues and guideline understanding. The vendor hired and compensated annotators at market rates in their respective countries. Annotators were required to be fluent in the source languages and native-level speakers in the translated languages. For rare language pairs, linguists were hired if professional graders were not available to ensure full language coverage. Two independent annotation rounds were conducted for each language pair. The segments in each round were distributed among different annotators. Each round was worked on by multiple annotators with the requirement that the same segment was annotated by two different annotators in the two rounds. Table~\ref{tab:lct_stats} provides statistics about the number of segments, the number of annotators, the number of translation systems and average segment length per language pair in our Long-Context Web MQM test set.

\subsection{Results on Long-Context Web MQM Test Set}
\label{sect:mqm_lct_results}

\subsubsection{Score and Error Count Evaluation}

\begin{figure*}[t]
\centering
\includegraphics[width=\textwidth]{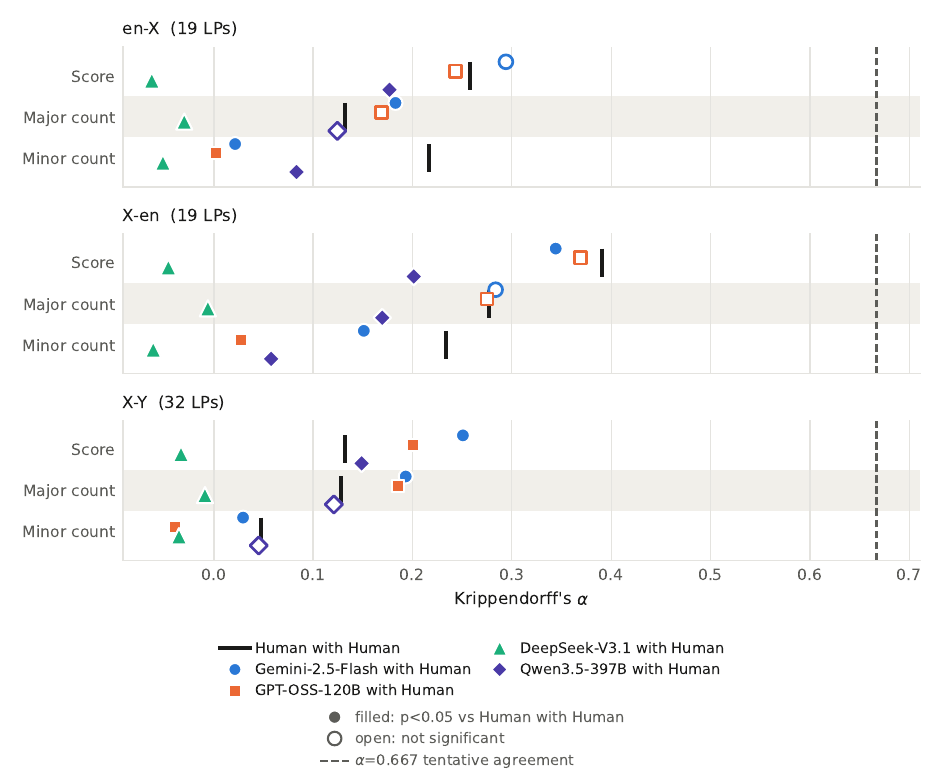}
\caption{Inter-annotator agreement on the Long-Context Web MQM test set, showing Krippendorff's $\alpha$ for MQM scores, major error counts, and minor error counts, averaged over the language pairs of each translation direction group (en-X, X-en, X-Y).}
\label{fig:alpha_by_direction_summary_MQM}
\end{figure*}

Figure~\ref{fig:alpha_by_direction_summary_MQM} shows MQM score, major and minor counts agreement averaged by language pair direction. For MQM score agreement for en-X language pairs, both \gemini{} and \gptoss-human average agreements are not significantly different from average human-human agreement while \deepseek{} and \qwen{}-human average agreements are significantly behind average human-human agreement. For X-en language pairs human-human average agreement is significantly larger than LLM-human average agreement for all LLMs except for \gptoss{}. For X-Y language pairs, all LLM-human average agreements are significantly larger than human-human average agreement for all LLMs except for \deepseek{}. It is worth noting that the human-human average agreement for X-Y language pairs is notably lower than human-human agreement for en-X and X-en language pair directions, which explains why LLM-human average agreement is significantly larger than human-human agreement for X-Y language pairs for most LLMs. Furthermore, all agreement levels are well below the 0.667 threshold for tentative agreement.

For major error counts, LLM-human average agreement outperforms or is on par with human-human average agreement for the three language pair directions for most LLMs. By contrast, LLM-human average agreement on minor error counts is significantly inferior to human-human average agreement for all language pair directions for most LLMs, which is similar to the results on the WMT23 test set in Section~\ref{sect:WMT23_MQM}.

  \begin{table}[ht]
\centering
\small
\resizebox{\columnwidth}{!}{
\begin{tabular}{lrcccc}
\toprule
\textbf{Direction} & \textbf{\#LPs} & \textbf{Gemini} & \textbf{GPT-OSS} & \textbf{DeepSeek} & \textbf{Qwen} \\
\midrule
en-X & 19 & 4/2 & 4/4 & 0/16 & 1/4 \\
X-en & 19 & 1/3 & 2/3 & 0/18 & 0/8 \\
X-Y & 32 & 20/2 & 19/4 & 3/21 & 9/6 \\
\midrule
All & 70 & 25/7 & 25/11 & 3/55 & 10/18 \\
\bottomrule
\end{tabular}}
\caption{Language pairs per translation direction whose LLM-human \kalpha{} on
MQM scores differs significantly from human-human agreement, on the
Long-Context Web MQM test set.
The first number counts language pairs where LLM-human \kalpha{} significantly exceeds human-human agreement, the second those where human-human significantly exceeds LLM-human. LPs with no significant difference in either direction appear in neither number.
Gemini = Gemini 2.5 Flash, GPT-OSS = GPT-OSS-120B, DeepSeek = DeepSeek-V3.1, Qwen = Qwen3.5-397B. Benjamini--Hochberg FDR applied separately to each LLM's comparisons, $p<0.05$.}
\label{tab:mqm_llm_exceeds_hh}
\end{table}

\begin{table}[t]
\centering
\footnotesize
\setlength{\tabcolsep}{2pt}
\begin{tabular}{lrrrrr}
\toprule
\textbf{Error category} & \textbf{H-H} & \textbf{Gem} & \textbf{GPT} & \textbf{DS} & \textbf{Qwen} \\
\midrule
Span overlap & .36\phantom{$^{*}$} & .33$^{*}$ & .24$^{*}$ & .23$^{*}$ & .28$^{*}$ \\
Character overlap & .28\phantom{$^{*}$} & .25$^{*}$ & .18$^{*}$ & .16$^{*}$ & .20$^{*}$ \\
Severity minor & .27\phantom{$^{*}$} & .19$^{*}$ & .15$^{*}$ & .15$^{*}$ & .18$^{*}$ \\
Mistranslation & .29\phantom{$^{*}$} & .30$^{*}$ & .19$^{*}$ & .12$^{*}$ & .21$^{*}$ \\
Severity major & .20\phantom{$^{*}$} & .24$^{*}$ & .16$^{*}$ & .10$^{*}$ & .17$^{*}$ \\
Awkward & .13\phantom{$^{*}$} & .06$^{*}$ & .07$^{*}$ & .03$^{*}$ & .07$^{*}$ \\
Grammar & .10\phantom{$^{*}$} & .09$^{*}$ & .07$^{*}$ & .06$^{*}$ & .09$^{*}$ \\
Omission & .24\phantom{$^{*}$} & .20$^{*}$ & .11$^{*}$ & .02$^{*}$ & .04$^{*}$ \\
Spelling & .17\phantom{$^{*}$} & .09$^{*}$ & .07$^{*}$ & .05$^{*}$ & .05$^{*}$ \\
Addition & .18\phantom{$^{*}$} & .13$^{*}$ & .12$^{*}$ & .07$^{*}$ & .09$^{*}$ \\
No error & .18\phantom{$^{*}$} & .15$^{*}$ & .14$^{*}$ & .00$^{*}$ & .09$^{*}$ \\
Punctuation & .18\phantom{$^{*}$} & .04$^{*}$ & .05$^{*}$ & .02$^{*}$ & .04$^{*}$ \\
Register & .05\phantom{$^{*}$} & .04$^{*}$ & .02$^{*}$ & .01$^{*}$ & .03$^{*}$ \\
Inconsistency & .13\phantom{$^{*}$} & .08$^{*}$ & .08$^{*}$ & .02$^{*}$ & .08$^{*}$ \\
Untranslated text & .21\phantom{$^{*}$} & .09$^{*}$ & .08$^{*}$ & .08$^{*}$ & .10$^{*}$ \\
Source error & .13\phantom{$^{*}$} & .06$^{*}$ & .00$^{*}$ & .00$^{*}$ & .00$^{*}$ \\
Wrong Language & .21\phantom{$^{*}$} & .13$^{*}$ & .09$^{*}$ & .00$^{*}$ & .06$^{*}$ \\
Name & .05\phantom{$^{*}$} & .00$^{*}$ & .00$^{*}$ & .00$^{*}$ & .02$^{*}$ \\
Currency & .23\phantom{$^{*}$} & .10$^{*}$ & .04$^{*}$ & .09$^{*}$ & .12$^{*}$ \\
Other & .04\phantom{$^{*}$} & .00$^{*}$ & .00$^{*}$ & .00$^{*}$ & .02$^{*}$ \\
Date & .25\phantom{$^{*}$} & .13$^{*}$ & .08$^{*}$ & .09$^{*}$ & .11$^{*}$ \\
Non translation & .05\phantom{$^{*}$} & .09$^{*}$ & .06\phantom{$^{*}$} & .08\phantom{$^{*}$} & .06\phantom{$^{*}$} \\
Time & .22\phantom{$^{*}$} & .07$^{*}$ & .05$^{*}$ & .06$^{*}$ & .07$^{*}$ \\
\bottomrule
\end{tabular}
\caption{Span-annotation F1 by error category on the Long-Context Web MQM test set, averaged over language pairs and ordered by decreasing category frequency. H-H is agreement between human annotation rounds; the remaining columns are each LLM's agreement with the human rounds. Gem = Gemini 2.5 Flash, GPT = GPT-OSS-120B, DS = DeepSeek-V3.1, Qwen = Qwen3.5-397B. $^{*}$ marks a difference from H-H that is statistically significant at $p<0.05$. Categories whose human-human baseline rests on fewer than 5 source segments are omitted, since no comparison against it is estimable: Address, Mismatched locale, Telephone.}
\label{tab:f1_by_categories_MQM}
\end{table}

Table~\ref{tab:mqm_llm_exceeds_hh} shows the number of language pairs per language pair direction where LLM-human agreement significantly exceeds human-human agreement versus the number of language pairs where human-human agreement significantly exceeds LLM-human agreement for MQM scores on the Long-Context Web MQM test set. For en-X and X-en directions, LLM-human and human-human agreements are not significantly different for most language pairs for all LLMs except for \deepseek{} which has an overall low agreement with humans. By contrast for X-Y language pairs, \gemini{} and \gptoss{}-human agreement outperform human-human agreement for most language pairs, which also stems from the low human-human agreement in this direction as demonstrated in Figure~\ref{fig:alpha_by_direction_summary_MQM}.

\begin{figure*}[p]
\centering
\includegraphics[width=\textwidth]{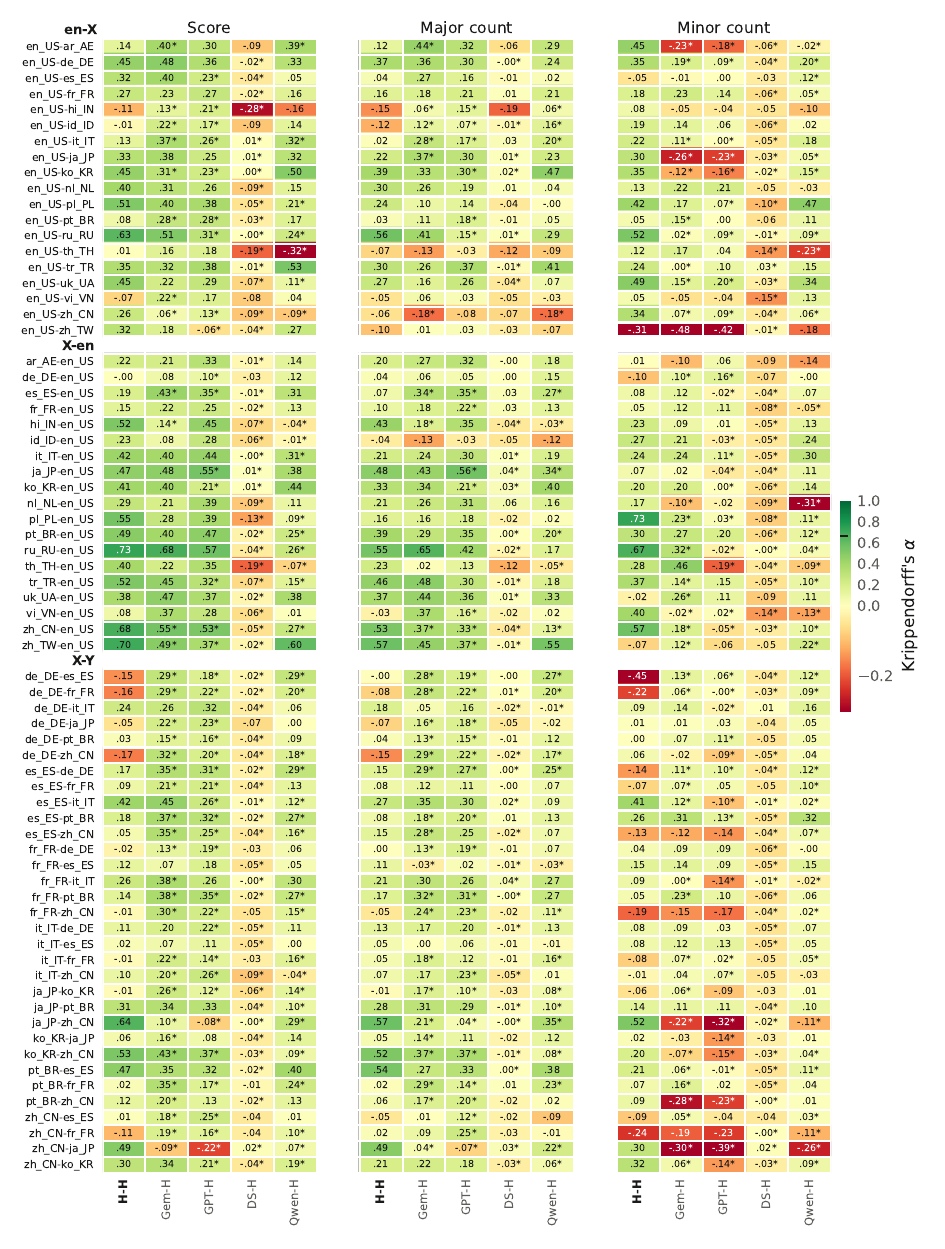}
\caption{Krippendorff's $\alpha$ agreement on MQM scores, major and minor error counts for each of the 70 language pairs of the Long-Context Web MQM test set, grouped by translation direction (en-X, X-en, X-Y). Within each group the columns are H-H (Human with Human), Gem-H (Gemini-2.5-Flash with Human), GPT-H (GPT-OSS-120B with Human), DS-H (DeepSeek-V3.1 with Human), Qwen-H (Qwen3.5-397B with Human). The dashed line on the colorbar marks $\alpha = 0.667$. $*$ marks cells where LLM-human agreement differs significantly from the human-human agreement of the same cell (paired cluster bootstrap on source segments, $p<0.05$).}
\label{fig:lct_alpha_heatmap}
\end{figure*}

 Figure~\ref{fig:lct_alpha_heatmap}
  presents MQM score, major error count and minor error count agreement results for individual language pairs within each direction. Both LLM-human and human-human agreement vary substantially between different language pairs. However, the overall agreement falls well below the \kalpha{} = 0.667 threshold for tentative conclusions for nearly all language pairs, with rare exceptions where human-human agreement reaches this threshold, including Russian-to-English (ru\_RU-en\_US), Simplified Chinese-to-English (zh\_CN-en\_US) and Traditional Chinese-to-English (zh\_TW-en\_US). Agreement is particularly low for Non-English pairs, which predominantly comprise low-resource language pairs. For humans this might indicate issues in the expertise and/or training of the human annotators for these language pairs. For LLMs, this is expected as lower-resource language pairs would be underrepresented in the LLM training data.

\subsubsection{Span Annotation Evaluation}
\label{sect:MQM_error_spans}

Table~\ref{tab:f1_by_categories_MQM} presents F1 scores
  for error category identification across annotator pairs, averaged over
  all language pairs. Similar to the WMT23 test set results in Section~\ref{sect:WMT23_MQM_error_spans}, agreement levels are low across all categories and annotator pairs with agreement reaching at most F1 score = 0.36. 

  Human-human agreement significantly exceeds LLM-human agreement for all span evaluation metrics except for mistranslation, major errors and non-translation where \gemini{}-human agreement significantly exceeds human-human agreement. All LLMs achieve low F1 for several low-frequency categories and all annotators have notably low agreement on identifying error-free segments (F1 score $\leq$ 0.18) consistent with the WMT23 test set results in Section~\ref{sect:WMT23_MQM_error_spans}.

\begin{figure*}[t]
\scriptsize
\begin{lstlisting}[basicstyle=\scriptsize\ttfamily,numbers=none,breaklines=true,frame=single,backgroundcolor=\color{gray!5}]
Translation Quality Evaluation

## Task Description
You are an expert translation quality evaluator.
You are given a text in the source language and a corresponding translation in the target language.
Your task is to evaluate the translation using Multidimensional Quality Metrics (MQM) according to instructions defined below.

## MQM Error Annotation Instructions
Please follow these guidelines to annotate all translation errors.

Please read the source and translation texts and annotate translation errors one by one, taking the context of the whole text into consideration. For each translation error please provide the following information:
Span: select the error span(s)
Category: select the error category from the error categories defined below.
Severity: select the error severity from the error severities defined below.

### Error Severity Levels
Severity    Description
Major       Errors that may confuse or mislead the reader due to significant change in meaning.
Minor       Errors that don't lead to loss of meaning and wouldn't confuse or mislead the reader but would be noticed, would decrease stylistic quality, fluency or clarity, or would make the content less appealing.


### Error Categories
Error Category                  Description
Accuracy/Addition               Translation includes information not present in the source
Accuracy/Omission               Translation is missing content from the source
Accuracy/Mistranslation         Translation does not accurately represent the source
Accuracy/Wrong Language         Translation includes segments in wrong target language. This doesn't include scenarios like names or quotes that are intended to be in another language.
Accuracy/Mismatched Locale      Translation includes segments in wrong locale variant.
Accuracy/Untranslated-text      Source text was left untranslated in the target when a translation is expected
Accuracy/Non-translation        Impossible to reliably characterize distinct translation errors because the translation is garbled or is unrelated to the source text
Fluency/Punctuation             Incorrect punctuation for locale or style
Fluency/Spelling                Incorrect spelling or capitalization
Fluency/Grammar                 Problems with grammar, other than orthography
Fluency/Register                Wrong grammatical register (eg, inappropriately informal pronouns).
Fluency/Inconsistency           Translation of the same word/phrase is inconsistent across the text
Style/Awkward                   Translation has inappropriate language style despite being accurate and grammatical
Locale convention/Address       Address format does not follow target locale conventions
Locale convention/Currency      Currency format does not follow target locale conventions
Locale convention/Date          Date format does not follow target locale conventions
Locale convention/Name          Name format does not follow target locale conventions
Locale convention/Telephone     Telephone number format does not follow target locale conventions
Locale convention/Time          Time format does not follow target locale conventions
Other/Source-error              An error in the source. (e.g. the text is not in the expected language, is garbled, nonsensical, has spelling errors or typos or non coherent)
Other/Other                     Any other issue in the source or translation

## Output Format
Return a JSON object with the following field:
- errors (list): list of translation errors found. Each error is an object with:
  - category (string): error category from the predefined list above (e.g. "Accuracy/Mistranslation")
  - severity (boolean): true for Major errors, false for Minor errors
  - other (string, optional): free-text description when category is "Other/Other"
  - spans (list): one or more error spans, each with:
    - start (integer): character index of the first character of the span (0-based, relative to the paragraph)
    - end (integer): character index of the last character of the span (inclusive, relative to the paragraph)
    - paraIndex (integer): index of the paragraph containing the span (0-based)
    - spanText (string): the actual text of the selected span
    - isSource (boolean): true if the span is on the source text, false if on the target
If there are no errors, return an empty list: {"errors": []}

## Translation to Evaluate
Source (<source_language>): <source_text>
Translation (<target_language>): <target_text>

## Your Evaluation

\end{lstlisting}
\caption{MQM LLM evaluation prompt template.}
\label{fig:mqm_prompt}
\end{figure*}

\begin{figure*}[t]
\scriptsize
\begin{lstlisting}[basicstyle=\scriptsize\ttfamily,numbers=none,breaklines=true,frame=single,backgroundcolor=\color{gray!5}]
Translation Quality Evaluation

## Task Description
You are an expert translation quality evaluator.
You are given a text in the source language and a corresponding translation in the target language.
Your task is to evaluate the translation using Error Span Annotation according to instructions defined below.

## Error Annotation Instructions
Select the part of translation where you have identified a translation error. Choose error severity:
- Minor errors: Style, grammar, word choice could be better or more natural.
- Major errors: The meaning is changed significantly and/or the part is really hard to understand.

Tips:
- Omitted content: If the translation is missing content from the source, set both start_i and end_i to "missing".
- Highlight the word or general area of the error (it doesn't need to be exact). Use multiple highlights for different errors.
- Pay particular attention to the language/variant being evaluated. If the translation is in the wrong language or language variant, mark it fully and assign it 0.
- Pay particular attention to translation consistency between texts across the whole document.
- If the translation contains additional text (e.g. "Here is the translation") or alternative secondary translation, mark it as a major error.

Rating Scale (from best to worst):

- Perfect [95-100]: The meaning of the translation is completely consistent with the source and reads like it was formulated by a native speaker.
- Retains core meaning with minor mistakes [67-94]: The translated sentence retains the core meaning of the source sentence, but may have some minor grammar mistakes, awkward formulations and/or might omit some insignificant parts.
- Preserves some meaning with major mistakes [6-66]: The translated sentence preserves some of the meaning of the source sentence but misses significant parts, or contains major errors.
- Nonsense or unrelated [0-5]: Completely nonsense translation that is not related to the source string.

Within each rating category, better translations should get higher scores.

## Output Format
Return a JSON object with the following fields:
- score (float): overall quality score in the range [0-100]
- errors (list): list of translation errors found. Each error is an object with:
  - start_i (integer or "missing"): character index of the first character of the error span (0-based), or "missing" if the error has no localized span (e.g. omission)
  - end_i (integer or "missing"): character index of the last character of the error span (inclusive), or "missing" if the error has no localized span
  - spanText (string): the actual text of the selected span, or null if missing
  - severity (boolean): true for Major errors, false for Minor errors
If there are no errors, return an empty list: {"score": <score>, "errors": []}

## Translation to Evaluate
Source (<source_language>): <source_text>
Translation (<target_language>): <target_text>

## Your Evaluation

\end{lstlisting}
\caption{ESA LLM evaluation prompt template adapted from ESA guidelines~\citep{kocmi-etal-2024-error}}
\label{fig:esa_prompt}
\end{figure*}

\FloatBarrier

\clearpage

{\footnotesize
\onecolumn
\begin{center}
\captionof{table}{Statistics for Long-Context MQM Test Sets. Ann shows the number of different annotators worked on each language pair in the two rounds, Segs shows the number of segments in each language pair, Sys shows the number of translation systems per language pair and Len shows average source segment length in words (\textsuperscript{w}) for space-using languages or characters (\textsuperscript{c}) for non-space languages.}
\label{tab:lct_stats}
\vspace{0.5em}
\begin{tabular}{lrrrr @{\hspace{1.5em}} lrrrr}
\toprule
\textbf{LP} & \textbf{Segs} & \textbf{Sys} & \textbf{Ann} & \textbf{Len} &
\textbf{LP} & \textbf{Segs} & \textbf{Sys} & \textbf{Ann} & \textbf{Len} \\
\midrule
ar\_AE-en\_US & 100 & 1 & 11 & 136\textsuperscript{w} & fr\_FR-es\_ES & 200 & 4 & 6 & 157\textsuperscript{w} \\
de\_DE-en\_US & 200 & 4 & 10 & 136\textsuperscript{w} & fr\_FR-it\_IT & 200 & 4 & 11 & 157\textsuperscript{w} \\
de\_DE-es\_ES & 200 & 4 & 13 & 136\textsuperscript{w} & fr\_FR-pt\_BR & 200 & 4 & 19 & 157\textsuperscript{w} \\
de\_DE-fr\_FR & 200 & 4 & 23 & 136\textsuperscript{w} & fr\_FR-zh\_CN & 200 & 4 & 15 & 157\textsuperscript{w} \\
de\_DE-it\_IT & 200 & 4 & 13 & 136\textsuperscript{w} & hi\_IN-en\_US & 100 & 1 & 11 & 142\textsuperscript{w} \\
de\_DE-ja\_JP & 200 & 4 & 16 & 136\textsuperscript{w} & id\_ID-en\_US & 100 & 1 & 10 & 133\textsuperscript{w} \\
de\_DE-pt\_BR & 200 & 4 & 23 & 136\textsuperscript{w} & it\_IT-de\_DE & 200 & 4 & 18 & 142\textsuperscript{w} \\
de\_DE-zh\_CN & 200 & 4 & 19 & 136\textsuperscript{w} & it\_IT-en\_US & 200 & 4 & 12 & 140\textsuperscript{w} \\
en\_US-ar\_AE & 100 & 1 & 8 & 138\textsuperscript{w} & it\_IT-es\_ES & 200 & 4 & 12 & 142\textsuperscript{w} \\
en\_US-de\_DE & 200 & 3 & 19 & 136\textsuperscript{w} & it\_IT-fr\_FR & 200 & 4 & 16 & 142\textsuperscript{w} \\
en\_US-es\_ES & 200 & 3 & 18 & 136\textsuperscript{w} & it\_IT-zh\_CN & 200 & 4 & 18 & 142\textsuperscript{w} \\
en\_US-fr\_FR & 200 & 3 & 17 & 136\textsuperscript{w} & ja\_JP-en\_US & 200 & 4 & 9 & 195\textsuperscript{c} \\
en\_US-hi\_IN & 100 & 1 & 11 & 138\textsuperscript{w} & ja\_JP-ko\_KR & 200 & 4 & 13 & 390\textsuperscript{c} \\
en\_US-id\_ID & 100 & 1 & 7 & 138\textsuperscript{w} & ja\_JP-pt\_BR & 200 & 4 & 15 & 390\textsuperscript{c} \\
en\_US-it\_IT & 200 & 3 & 22 & 136\textsuperscript{w} & ja\_JP-zh\_CN & 200 & 4 & 9 & 390\textsuperscript{c} \\
en\_US-ja\_JP & 200 & 3 & 10 & 136\textsuperscript{w} & ko\_KR-en\_US & 200 & 4 & 10 & 141\textsuperscript{w} \\
en\_US-ko\_KR & 200 & 3 & 5 & 136\textsuperscript{w} & ko\_KR-ja\_JP & 200 & 4 & 17 & 99\textsuperscript{w} \\
en\_US-nl\_NL & 100 & 1 & 6 & 138\textsuperscript{w} & ko\_KR-zh\_CN & 200 & 4 & 18 & 99\textsuperscript{w} \\
en\_US-pl\_PL & 100 & 1 & 6 & 138\textsuperscript{w} & nl\_NL-en\_US & 100 & 1 & 5 & 137\textsuperscript{w} \\
en\_US-pt\_BR & 200 & 3 & 19 & 136\textsuperscript{w} & pl\_PL-en\_US & 100 & 1 & 5 & 141\textsuperscript{w} \\
en\_US-ru\_RU & 100 & 1 & 4 & 138\textsuperscript{w} & pt\_BR-en\_US & 200 & 4 & 19 & 141\textsuperscript{w} \\
en\_US-th\_TH & 100 & 1 & 3 & 138\textsuperscript{w} & pt\_BR-es\_ES & 200 & 4 & 17 & 146\textsuperscript{w} \\
en\_US-tr\_TR & 100 & 1 & 8 & 138\textsuperscript{w} & pt\_BR-fr\_FR & 200 & 4 & 15 & 146\textsuperscript{w} \\
en\_US-uk\_UA & 100 & 1 & 5 & 138\textsuperscript{w} & pt\_BR-zh\_CN & 200 & 4 & 19 & 146\textsuperscript{w} \\
en\_US-vi\_VN & 100 & 1 & 6 & 138\textsuperscript{w} & ru\_RU-en\_US & 100 & 1 & 6 & 139\textsuperscript{w} \\
en\_US-zh\_CN & 200 & 3 & 8 & 136\textsuperscript{w} & th\_TH-en\_US & 50 & 1 & 3 & 884\textsuperscript{c} \\
en\_US-zh\_TW & 100 & 1 & 10 & 138\textsuperscript{w} & tr\_TR-en\_US & 100 & 1 & 8 & 138\textsuperscript{w} \\
es\_ES-de\_DE & 200 & 4 & 18 & 150\textsuperscript{w} & uk\_UA-en\_US & 100 & 1 & 9 & 137\textsuperscript{w} \\
es\_ES-en\_US & 200 & 4 & 21 & 136\textsuperscript{w} & vi\_VN-en\_US & 100 & 1 & 5 & 144\textsuperscript{w} \\
es\_ES-fr\_FR & 200 & 4 & 18 & 150\textsuperscript{w} & zh\_CN-en\_US & 200 & 4 & 8 & 215\textsuperscript{c} \\
es\_ES-it\_IT & 200 & 4 & 21 & 150\textsuperscript{w} & zh\_CN-es\_ES & 200 & 4 & 7 & 276\textsuperscript{c} \\
es\_ES-pt\_BR & 200 & 4 & 20 & 150\textsuperscript{w} & zh\_CN-fr\_FR & 200 & 4 & 17 & 276\textsuperscript{c} \\
es\_ES-zh\_CN & 200 & 4 & 8 & 150\textsuperscript{w} & zh\_CN-ja\_JP & 200 & 4 & 7 & 276\textsuperscript{c} \\
fr\_FR-de\_DE & 200 & 4 & 26 & 157\textsuperscript{w} & zh\_CN-ko\_KR & 200 & 4 & 13 & 276\textsuperscript{c} \\
fr\_FR-en\_US & 200 & 4 & 17 & 138\textsuperscript{w} & zh\_TW-en\_US & 100 & 1 & 6 & 202\textsuperscript{c} \\
\bottomrule
\end{tabular}
\end{center}
}

\begin{center}
\captionof{table}{Statistics for the WMT23 MQM test set. Len shows average source segment length in words (\textsuperscript{w}) for space-using languages or characters (\textsuperscript{c}) for non-space languages.}
\label{tab:wmt23_mqm_stats}
\vspace{0.5em}
{\footnotesize
\begin{tabular}{lrrrr}
\toprule
\textbf{LP} & \textbf{Segs} & \textbf{Sys} & \textbf{Ann} & \textbf{Len} \\
\midrule
en\_US-de\_DE & 190 & 13 & 10 & 144\textsuperscript{w} \\
zh\_CN-en\_US & 179 & 16 & 8 & 264\textsuperscript{c} \\
\bottomrule
\end{tabular}
}
\end{center}

\FloatBarrier

\applefootnote{ \textcolor{textgray}{\sffamily Apple and the Apple logo are trademarks of Apple Inc., registered in the U.S. and other countries and regions.}}

\end{document}